# Title: A Soft Robotic Demonstration in the Stratosphere

**Authors:**

Codrin Tugui[1,2]†, Tirth Thakar[1]†, Anatol Gogoj[1]†, Alexander White,[1] Ang "Leo" Li,[1,3] Alexander Yin[1], Edward Pomianek[1], Mihai Duduta[1]*

**Affiliations:**

[1]University of Connecticut, School of Mechanical, Aerospace, and Manufacturing Engineering, Storrs, Connecticut USA

[2]Institute of Macromolecular Chemistry Petru Poni; Iași, Romania

3University of Toronto, Department of Mechanical and Industrial Engineering, Toronto, Ontario, Canada

*Corresponding author. Email: mihai.duduta@uconn.edu

†These authors contributed equally to this work.

**Abstract:** Machines designed for operation in Space, as well as other extreme environments, need to be both resilient and adaptable when mission parameters change. Soft robots offer advantages in adaptability, but most lack resilience to the pressure and temperature extremes found as close as the Stratosphere. Dielectric elastomer actuators overcome some of those limitations when built as solid state compliant capacitors capable of converting electrical energy into mechanical work, but the elastomer resilience limits the device's operating window. Here we present a crosslinking mechanism for silicone elastomers under ultraviolet light using trimethyl(methylcyclopentadienyl)platinum(IV) as a catalyst to react hydrosilane to vinyl groups. The formation of carbon-carbon bonds enables fast processing under UV light and exceptional electro-mechanical performance in dielectric elastomer actuators. The material resilience advantage is demonstrated in controlled experiments at -40° and 120° C, as well as near vacuum, in comparison with state-of-the-art acrylic and silicone chemistries. Fully autonomous systems controlling grippers made with the novel silicone were integrated into payloads for high altitude balloon testing. Two stratospheric balloon missions were carried out and demonstrated DEAs as a viable soft robotic technology under space-like conditions (as high as 23.6 km elevation, at <0.05 atm and -55° C). The combinations of chemical building blocks and catalyst can be further expanded to address other challenges for silicones, including adhesion and additive manufacturing.

**One-Sentence Summary:** Novel crosslinking mechanism enables silicone-based dielectric elastomer to demonstrate actuation in the Stratosphere at low temperature and pressure.

## Introduction

Soft robots are a subclass of robotic systems that are highly advantageous over traditional robotic systems that are rigid and nonconformal in nature. Due to this inherent adaptability, soft robots have been proposed for carrying out tasks in unconstrained environments like Space, the Deep Sea, and other challenging locations. The adaptability of soft robots can be hindered by the driving mechanism by which they achieve actuation *(1)*. These include Shape Memory Alloys (SMA) *(2)*, fluid driven actuators *(3)*, electroactive polymers (EAPs) *(4)*, electrical and magnetic *(5)*, light *(6)*, sound *(7)* and chemical *(8)* actuators. In the past, there have been a few notable soft robots demonstrated to operate in a variety of extreme environments (Figure 1A) such as a pneumatically operated search and rescue robot operating in an open flame environment that reached nearly 280°C *(9)*, a boa inspired bellows-type gripper actuator for reef sample collection *(10)*, and an electroactive exploration robot that survived at the bottom of the Mariana Trench *(11)*. However, Space and extra-terrestrial environments exhibit large variations in temperature (from –70° to 120° C in the span of several hours in Low Earth Orbit) and pressure which limit the viability of most listed technologies. Focusing specifically on dielectric elastomer actuators (DEAs), this works establishes a novel crosslinking mechanism, and evaluates the resulting materials' suitability for applications in Space missions.

DEAs are a type of soft robotic actuator that relies on the principle of electrostatics to generate actuation. DEAs in their most basic form comprise an elastomer layer sandwiched between two mechanically compliant electrodes *(12)*. As an electric field gets applied to the electrodes, the Coulombic attraction between them causes the elastomer layer to be compressed and undergo lateral expansion due to the incompressibility of the material. The resultant force produced through this method of actuation is low and in the ranges of 1-10mN *(13)* and thus the applications of DEAs are very limited in this form. However, stacking these building blocks of electrode-elastomer-electrode into a multilayered configuration (Figure 1B), allows us to obtain higher forces (>10 N) and displacements (>1 cm), which corresponds to specific energy on par with natural muscles (>20 J/kg) *(14)*. The solid-state construction allows DEAs to be pressure and temperature tolerant within the limits of the constitutive materials. In space-like environments DEAs can be adapted to carry out tasks including locomotion in unstructured environments *(15)*, surface exploration *(16)*, sample collection *(17)*, and capturing space debris *(18)*.

To date, two major types of polymer chemistries have been used for the construction of multilayer DEAs: ultraviolet (UV) light curable acrylics *(19)* and thermally curable silicones *(20)*. The trade-offs between the two are well established: acrylics offer faster processing times under UV light, a higher dielectric constant *(21)*, a higher specific energy, and better adhesion for rapid integration into robotic components. Meanwhile silicones offer improved cycle life *(22)* and higher specific power, enabling unique applications such as flapping wing micro robots *(23)*. The limitations of silicones, in contrast with acrylics as building blocks for DEAs, are longer processing times, poor adhesion to non-silicone substrates, and lower dielectric constant. Recent advances have made some improvements, including increasing the dielectric constant by blending commercial silicones *(24)*, and leveraging thiol-ene groups for increasing strain *(25)*. Even for acrylics, the focus is on maximizing the dielectric constant *(26)*, while the cycle life remains poor (only 10,000 cycles reported) and the glass transition temperature is too high for the low temperatures of space.

In this work, we present a crosslinking mechanism reacting hydrosilane to vinyl groups in the presence of trimethyl(methylcyclopentadienyl)platinum(IV) catalyst under ultraviolet light to produce UV curable resilient silicone elastomers (UV-RSE). The reaction parameters are presented in the first part of the results section, with the rest of the paper aimed at establishing DEAs and these materials as suitable building blocks for space applications. The UV-RSEs show a significant improvement over state-of-the-art dielectric elastomer actuator materials and can be employed in harsh environment applications. To prove this claim, we carried out experiments in controlled environments at pressure and temperature extremes on single-layer and multilayer devices (Figure 1D). The UV-RSE was tested in comparison to a urethane acrylate-based elastomer CN9018 *(19)*, and a commercial silicone elastomer, Dragonskin-20. As a final demonstration, we tested the materials as simple free moving grippers powered autonomously during two high altitude flying missions to the Stratosphere (Figure 1C). In this near Space environment, the materials demonstrated exceptional performance at high elevation (23.6 km) and low temperature (-55° C) and pressure (0.05 atm). The robust performance of the entire payload demonstrates that solid state dielectric elastomer actuators are ready for Space applications, that science in the Stratosphere is accessible, and that harsh environment can serve as proving ground for novel robotic technologies. Beyond actuation, the chemical building blocks can be further tuned to address remaining challenges in silicone polymers, including adhesion and additive manufacturing.

The remainder of the paper is organized as follows. In the section "Results", we present the crosslinking mechanism, electro-mechanical performance of the elastomers under simulated harsh conditions, and the results of testing a simple robot gripper in the Stratosphere. The "Conclusion" section presents the implications and conclusions of our work and highlights future improvements that can be made to push soft robots towards valuable applications in extreme and harsh environments.

## Results

### *Chemical Crosslinking Mechanism*

The reactivity of hydrosilane groups towards unsaturated carbon – carbon bonds has been well established in literature, with multiple groups demonstrating these types of reaction driven thermally *(27)*. Both telechelic vinyl functional polydimethyl siloxane (PDMS) and monofunctional vinyl PDMS have been reacted with hydride groups at different locations on PDMS chains to produce stretchable or soft silicone elastomers *(28)*. For photocurable silicones, different strategies have been employed, crosslinking by reacting either acrylate esters *(29)*, epoxides *(30)* or thiol groups *(31)*. While effective, each of these methods has limitations, and adds significant complexity to the preparation of the starting materials. In contrast, the reactivity of hydrosilanes with unsaturated carbon – carbon bonds under ultraviolet light has been established in the processing of silicon wafers *(32)*. Building on this knowledge, some effort has gone into developing catalysts for UV curing of silicones, without a focus on maximizing electro-mechanical performance of the elastomers *(33)*.

The building blocks we used for this crosslinking mechanism are shown in Figure 1E, and consist of telechelic vinyl terminated PDMS (Block A), hydrosilanes (e.g. methylhydrosiloxane, co-polymerized with dimethyl siloxane) (Block B), and a UV light induced photocatalyst (e.g. for this work only trimethyl(methylcyclopentadienyl) platinum (IV)) (Block D). Interestingly, commercial products exist which are mixtures of the key components: specifically Part B of Elastosil P7670 made by Wacker Chemie AG contains both the telechelic vinyl terminated PDMS and hydrosilane functionality (Block C), which allows the prepolymer mixture sold as Part B to cure under ultraviolet light in the presence of the photocatalyst. The catalyst is dissolved in 1,3-dioxolane and is used in small enough quantities that the evaporation of the carrier solvent does not appreciably impact the properties of the final elastomeric film. Unlike acrylics which need surrounding oxygen to be removed for UV curing to occur, the PDMS prepolymer reacts rapidly when illuminated at 50 $mW/cm^2$ for 3 minutes, with minimal changes if the curing time is extended for as much as 20 minutes.

We explored the parameter space by varying the composition of the prepolymer by adding either Block A or Block B only to the commercial mixture that contained both functionalities (Block C) as shown in Table 1. Across the small variations, we found the mechanical properties were slightly changed by increasing the amount of photocatalyst by a factor of 10x. Tripling the amount of platinum catalyst (Block C) from 100 to 300 ppm caused the Young's modulus to more than double (from 1.5 to 3.3 MPa), while the tensile elongation decreased from 650 to 500%. Further increases in the catalyst amount, up to 1000 ppm did not significantly impact either the modulus of elongation. Adding Block A caused an expected change, decreasing the modulus to 2.3 MPa and increasing the elongation to 900% before failure. Adding Block B worked the opposite way, increasing the modulus to 4.1-4.4 MPa, and reducing the elongation to less than 300% before onset of failure. The results are consistent with the expected reactivity of the starting materials, while all elastomeric films were nearly 100% transparent and cured within 3-5 minutes. The electrical properties of the films made with varying amounts of catalyst are summarized in Figure S7, while the data compiled in Table 1 is extracted from Figures S8 and S9.

The crosslinking mechanism can be adapted towards other applications by varying either of the starting blocks towards other goals. For example, using a telechelic hydride functionalized PDMS (such as DMS-H25 from Gelest) can drastically extend the chain length, allowing for ultrahigh elongation to break. Alternatively, using a multi-hydride PDMS (such as HMS-064 from the same company) can lead to a bottlebrush structure with ultralow modulus *(34)*. For the goals set out in this work, i.e. to develop a UV-curable material that retained electro-mechanical performance at pressure and temperature extremes, combinations of Blocks A, B and C were sufficient and outperformed state-of-the-art elastomers, including both commercial silicones and custom made acrylic urethane mixtures. For the next section, the focus was on characterizing and understanding the elastomer and electrode fatigue in simulated conditions that mimic operation in the Stratosphere, including temperature changes from –55° to 120° C, and pressure decreases as low as 0.05 atm.

### ***Controlled Environment Testing***

Materials and devices that need to operate in Space are tested rigorously before deployment to prevent dangerous and expensive failures. While the effects of space environments on silicone materials have been well documented *(35)*, we aimed to understand the impact of electro-mechanical actuation on elastomers under similar conditions to Low Earth Orbit (LEO). One brief exploration of the potential for DEAs for space manipulation only focused on the kinematics of the system, without addressing the viability of the materials *(36)*. For this section, we set out to test single layer DEAs at low (-40° C) and high temperatures (120° C), as well as low pressure (5 kPa), considering the environmental factors as independent variables. We studied both the change in actuation strain as a function of environmental conditions, as well as the operational cycle life, targeting 10,000 actuation cycles as a goal for all elastomer types.

#### *Actuation Strain Variation with Temperature*

We gauged the effects of temperature on the resulting actuation strain in our single-layer DEAs by subjecting them to different temperature conditions (-40°, 20° and 120° C) and incrementally increasing the electric field to the electrodes and measuring the actuation strain via a high-definition camera and measurement software. This was then repeated for a total of 10 samples for each temperature condition at for each material. The results for this experiment can be seen in Figure 1D, where the dashed lines show the median dielectric breakdown strength of each material at each temperature condition. Following the governing Maxwell's stress equation, the strain depends approximately on the square of the applied electric field, however the rate of increase in strain is not constant across all temperatures. We observed that the acrylic based elastomer CN9018 exhibits relatively low actuation strain when exposed to cold temperatures as compared to ambient and hot environments The behavior is expected as the glass transition temperature for CN9018 is -55°C, meaning its stiffness increases drastically, making it harder to compress. Moreover, the actuation strain for CN9018 decreases at higher temperatures compared to ambient condition at the same electric field, confirming the poor viability as a material for Space applications.

In contrast, we found both silicone materials to perform well at low temperature, an expected finding given the low glass transition temperature of the polydimethylsiloxane backbone (near -120°C). The UV-RSE shows the least variation between temperatures at equivalent fields, with slightly higher actuation strain at lower temperatures. In contrast, Dragonskin-20 showed a slight decrease in actuation strain under cold conditions compared to its ambient and hot condition counterpart at the same electric field. The main drawback of the Dragonskin-20 material was the low breakdown strength at high temperature (Fig. S12), which had a significant impact on its cycle life, as described below. Across the three material options, the UV-RSE had the smallest variation in area strain across temperatures, with the smallest variation in breakdown field, making it the most reliable option for a mission to the Stratosphere.

*Strain Variation with Pressure*

The low-pressure environment of Space drastically limits the options for actuation in soft robotic systems, as any fluid driven approaches *(37, 38)* can be rendered inoperative by a fluid leak. Under low-pressure conditions, many elastomers are affected by outgassing which can cause macroscopic changes in the properties of the material *(39)*. To date, studies on the pressure impact on DEAs have focused on their potential applications as vibration control devices *(40)*. To quantify the effects of low-pressures on our single-layer DEAs, we used a vacuum chamber that achieved an absolute pressure of 5 kPa which equates to less than 5% of atmospheric pressure. This pressure is in accordance with that found in the near-space and stratosphere regions, and similar to established environmental testing approaches *(41)*. Following the temperature testing, we fabricated and tested 10 single-layer devices for each material (i.e. acrylic CN9018, Dragonskin-20 and the UV-RSE). We made one important change to each deivce to circumvent the effects of vacuum arcs, which form when high electric fields are exposed to low-pressure environments *(42, 43)*. We coated each device in a thin layer of elastomer to encapsulate the exposed electrodes. We first measured the actuation strain on each of the samples at atmospheric pressure at a given electric field and then repeated using the same sample and electric field but now at a low-pressure condition. In this manner, we directly compared the changes in the level of actuation due to a change in pressure for that specific device.

The results, summarized in Figure 2C, show that both silicone-based elastomers are minimally impacted by the change in pressure with a deviation of ±10% in actuation strain when exposed to low-pressure conditions. Meanwhile the CN9018 samples exhibit a considerable deviation of nearly 30% in some instances, with the average drop in actuation being around 10% compared to atmospheric pressure. The poor performance of the acrylic is likely due to a combination of factors, including both outgassing and migration of mobile species in the elastomer. The acrylic is made from eigth different components, including a volatile solvent, and incomplete curing of the entire matrix can have a detrimental impact on the device's electro-mechanical performance. In contrast, the silicone prepolymer mixtures were either completely converted into solid films or used minimal amounts of solvent for the UV photocatalyst that was readily removed during the initial film curing step.

*Cycle Life Variation with Temperature*

Dielectric elastomer actuators show a complex degradation mechanism that depends on multiple operational and environmental factors *(44-46)*. To capture the combined impact of electrical and mechanical stress, we performed repeated actuation cycles at the same set temperatures, up to a goal of 10,000 cycles per device. We used this set of experiments as the final selection threshold for picking the elastomer material to be made into multilayer actuators for a Stratosphere test. As before, we made 10 single-layer devices using each elastomer material for each temperature condition. To make the head-to-head comparison between elastomers more reliable, we tested each material to 50% of the median dielectric breakdown established for that material, at that temperature (Figure S12). The value of 50% is derived from standard mechanical fatigue testing where the load applied to a sample is 50% of its tensile strength, and established protocols for DEAs *(47, 48)*. Each sample was first actuated once to obtain the baseline actuation strain and then actuated continuously with strain measurements being taken every $10^3$ cycles until the device no longer showed any measurable strain, experienced breakdown or achieved $10^4$ cycles. This allowed us to capture two valuable sets of information, the cycle life of each DEA sample, and its change in actuation strain for different temperature conditions. The overview of this testing and data collection method is shown in Fig S6.

Figure 2D shows that the UV-RSE survived the entire $10^4$ cycles of actuation across all samples and temperature conditions without experiencing dielectric breakdown. Dragonskin-20 devices under ambient temperatures also survived all $10^4$ cycles, but showed significant degradation at both low and high temperatures. We note that the Dragonskin material was tested to different fields at the different temperatures, due to its change in dielectric strength, which may add some variability to the data. Following the trends shown during the initial temperature testing, the acrylic-based elastomer CN9018 performed poorly in cold environments and exhibited a poor cycle life of only 100 cycles. The standout material is the UV-RSE, which has some degradation in the strain reached at the same field between the first and the last cycle in the $10^4$ cycle sweep, but overall survives the repeated testing at all temperatures.

***Stratosphere Demonstration***

For soft robots to be adopted broadly, they need to show capability outside of the research laboratory, in realistic settings for valuable applications. While multiple soft robotic systems have been demonstrated underwater *(49, 50),* to date there have been no demonstrations of soft robots in near Space conditions, Given the simple operation, and material resilience, DEAs made with the novel UV-RSE are well suited for operation in Space and near-Space environments. We chose to pursue a soft robotic demonstration in the Stratosphere because it is a well-established test environment for near Space conditions *(51)*, and more affordable due to the high-altitude balloon launch compared to a rocket launch. Conducting a robotic experiment during a high-altitude balloon flight into the stratosphere constitutes a significant scientific achievement in its own right, as it requires reliable autonomous operation of a tightly mass- and power-constrained payload under extreme conditions of low pressure, low temperature, and limited communication. Successfully completing the test demonstrates not only the soft actuator's functional robustness, but also the maturation of system-level design, integration, and validation strategies that are directly relevant to other inaccessible, high-risk environments.

*High Altitude Balloon Flight Experiment*

The soft actuator payload was carried on a high-altitude balloon, launched on December 16, 2025, from Ashland, Nebraska (41.0162°N 96.3182°W) by StratoStar. The balloon ascended at a maximum rate of 29 m/s into the Earth's stratosphere before bursting at a GPS altitude of 23,640 meters. During the 91-minute flight, the balloon traveled eastward with a total down-range distance of 85 km before landing in Coburn, Iowa (40.9104°N 95.3082°W). The payload was carried by a high altitude hydrogen ballon as described in the Supplemental Information.

The conditions experienced during the flight exceeded the low pressure and low temperature parameters used in the controlled validation experiments for the three elastomers, as outlined in Figure 3E and 3F. The ambient temperature of the stratospheric flight dropped to approximately –50°C while the pressure dropped to approximately 40.13 hPa (4013 Pa), or just under 4% of sea level atmospheric pressure. A GPS trace of the balloon flight is shown in Fig. 3C rendered in Google Earth Pro. Figure 3G shows the three axis accelerations experienced by the system, both recorded by the central payload and sensors adjacent to the actuators in our payload.

The payload construction included single and multilayer actuators, two independent high voltage power sources, environmental sensors and cameras to record the experiment. Figure 3A shows a diagram of the payload construction, and Figure 3B shows how the multilayer and single layer actuators were arranged in a custom modular frame. The electronics topology is described in detail in Figure S5, and was used to drive the single and multi-layer actuators alternatively. Two identical circuits and sets of actuators were built and installed for redundancy, with only the orientation of the setup being varied between the two sets. One set of multilayer actuators was aligned parallel with gravity, and the other perpendicular to gravity. The actuators were operated at 1.8 kV, which was sufficient for appreciable bending deformation, and low enough to eliminate the risk of dielectric breakdown which would have jeopardized the entire mission and payload. Because the multilayer and single layer actuators were driven with similar magnitude voltages, the single layer deformation was too small to be captured by the on board cameras at their resolution.

As shown in the pressure and temperature charts, the robot fully recorded its conditions up to around 12 km, before experiencing data loss. The cause of the data loss is likely a power surge to the controllers, which stopped the operation of the entire payload. However, as shown in Figure 4A, the UV-RSE actuators were fully capable of bending near the point where data loss occurred. The point corresponds to approximately 12 km elevation, where our own sensors recorded a temperature of -50° C and a pressure of 4 kPa. The difference between the temperature recorded by our sensors and the central payload is due to the fact that the central payload was enclosed, while our setup was exposed to the external environment. At the same pressure and temperatures, the CN9018 actuators attached to the second frame did not show any movement, an expected result given the elastomer behavior in controlled environment testing. The range of motion of the actuators in flight is recorded in Supplemental Video S1.

After the payload was recovered in Iowa, it was shipped back to our research laboratory for survivability testing. We tested all actuators on board and found all to be operational, with the exception of one single layer actuator which was damaged during landing. Figure D shows the range of motion of the recovered actuators, which have similar performance to what was recorded before flight during a practice run. The range of motion comparison can be found in Supplemental Videos S2 and S3 for the different actuator pairs. Given that the actuators were operational at the lowest temperature reached, and after landing, we consider the mission to be a complete success, despite the data loss. In contrast, the still image from Figure 1C was recorded during the first mission attempt, described in more detail in the Supplemental Information. During the first flight, the entire power module failed, and no actuation was observed at altitude. Once those actuators were recovered, they were still fully operational, allowing us to use the first mission as a survivability test.

## Conclusion

The need for resilient and adaptable robotic systems in Space and other harsh environments is growing, but few soft material technologies are suitable for these extreme conditions. Compared to other soft active materials, dielectric elastomer actuators have multiple advantages, including their high power and energy densities, solid-state construction, and simple electro-mechanical transduction mechanism. However, the elastomer material is the critical component, as the pressure and temperature extremes can drastically reduce the range of motion and actuator cycle life. As shown in this work, state-of-the-art acrylic materials exhibit limited performance at both low and high temperatures, and are unsuitable for Space-like conditions. Silicone elastomers have the benefit of a low glass transition temperature in the siloxane backbone, and can operate at low temperatures but some commercial materials, like Dragonskin-20, are limited at high temperature.

To overcome these challenges, we have established a faster crosslinking approach for hydrosilanes and vinyl groups that overcomes the typical photocuring limitations of silicones. The platinum catalyst allows for fast curing under ultraviolet light without the shrinkage due to acrylic groups, the unpleasant smell of thiol functionalities, and the high sensitivity to moisture of epoxy moieties. The material developed for the Stratosphere application has specific properties aimed at robust electro-mechanical performance in a relatively stiff elastomer, and is not optimized towards any specific power or energy density goal. The simple building blocks outlined here can be further modified to address other limitations of silicones, including adhesion to other substrates and reliable additive manufacturing, such as 3D printing, including multi-material fabrication. This first of its kind demonstration of a soft robot in the Stratosphere shows both the resilience of DEAs as soft actuators, as well as the maturity of the power electronics required to drive the system.

## Materials and Methods

### *Elastomer Film Fabrication and Testing*

The CN9018 elastomer was made following the established procedure *(19),* while the Dragonskin-20 was mixed following the supplier instructions (i.e. a 1:1 blend of Part A and Part B, mixed with a Flacktek planetary centrifugal mixer). The UV-RSE was made with Part B of Elastosil P 7670 (Wacker, Munich, Germany) mixed with different components. These include: Block A (Vinyl-terminated polydimethylsiloxane, average molecular weight ($M_w$) $M_w$ = 43000 grams per mole (g/mole) purchased as DMS-V33). Block B was Polydimethylsiloxane hydride, average molecular weight ($M_w$) = 25,000 grams per mole g/mol, and 4-6 mole percent (mole%) methylhydrosiloxane units, purchased as HMS-053 from Gelest. The catalyst was Trimethyl(methylcyclopentadienyl)platinum(IV)(IR light-induced photocatalyst), 10 weight percent (wt%) in 1,3-dioxolane, both purchased from Milipore Sigma. Elastomer layers were dispensed using a thin film blade coater (Elcometer 4340, Manchester, UK), and cured under UV light at 365 nm, or in an oven. The tensile testing followed ASTM standard D638 type one, in an Instron 3400 with nitrogen cooling for low temperature measurements.

***Temperature Testing Setup***

To create a cold environment for experiments that required cold extreme temperatures while also allowing a clear view of the actuator for measurement, an open system that reached a low steady state temperature was assembled. The assembly as shown in Figure S2 consists of dry ice submerged in a tray full of acetone. Also in the tray was a flat bottom crystallizing plate filled with silicone oil. As the acetone and dry ice mixture encountered plate with the silicone oil it solidified and reached the temperature of the bath. This created a stable platform on which the device could be placed upon and then covered, allowing the device’s temperature to also drop down to -40°C. Before each test iteration was carried out on a new sample, the surface temperature of the actuator sample was verified using a Type K digital thermocouple.

For tests involving hot environmental temperatures, a laboratory grade convection oven was used. The oven was set to 120°C at the start of every experiment and was confirmed using the Type K digital thermocouple. Once confirmed, the single-layer devices were fixed to the window of the oven using a double-sided adhesive to hold the actuator perpendicular to the camera for measurements. The assembly of this setup can be seen in Figure S1. Once the temperature of the actuator stabilized to the temperature of its hot environment, the test procedure for each respective experiment was carried out.

***Pressure Testing Setup***

The vacuum testing setup shown in Fig. S3 involved the use of a vacuum chamber that maintained an absolute pressure of 5000 Pa which equates to less than 5% of atmospheric pressure. This was a suitable environment to test the feasibility of our devices under low pressure / near-vacuum environments. Due to the o-rings used in the vacuum chamber it was necessary to include a high-voltage controllable circuit to provide power to the actuator directly. The circuit consisted of a 7.4V Lithium Polymer battery, and an Arduino Nano connected to an HVM high voltage DC-DC converter. The system was powered with an IR remote connected to the Arduino.

***Launch vehicle platform***

The overall structure of the high-altitude balloon (HAB) system is shown in Figure S11 including a balloon at the top, a radar retroreflector, a tracking system, a recovery parachute, and a biological payload. The balloon was targeted to reach an apogee 23.6 km above sea level in the stratosphere. The balloon used in the mission was specified as a 2000-series latex balloon, initially inflated using hydrogen gas, providing the required lifting force for the ascent.

Acknowledgments: The authors thank Jason Krueger and Michael Sibbernsen of StratoStar for guidance on payload integration for the high-altitude balloon launch.

**Funding:**

MidAtlantic Hub of the Air Force Research Laboratory - Prototyping Award (Fall 2023).
National Institute for Undersea Vehicle Technology – SEED 61 grant.
DEVCOM Ground Vehicles Systems Center – through the Digital Design Research, Analysis, and Manufacturing Center at the University of Connecticut.
NASA Connecticut Space Grant Graduate Fellowships: AW, EP, AMG.
Pratt & Whitney Institute for Advanced Systems Engineering Fellowships: TT, EP.
Launch support from the School of Mechanical, Aerospace and Manufacturing Engineering at the University of Connecticut.

**Author contributions:**

Conceptualization: MD, CD, TT, AMG
Methodology: CT, TT, AMG, AL, AW
Investigation: CT, TT, AMG, AL, AW, AY, EP
Visualization: CT, TT, AMG
Funding acquisition: MD
Project administration: MD, TT, AMG
Supervision: MD
Writing – original draft: CT, TT, AMG, MD
Writing – review & editing: MD

**Competing interests:** Authors declare that they have no competing interests.

**Data and materials availability:** All data, code, and materials used in the analysis must be available in some form to any researcher for purposes of reproducing or extending the analysis. Include a note explaining any restrictions on materials, such as materials transfer agreements (MTAs). Note accession numbers to any data relating to the paper and deposited in a public database; include a brief description of the data set or model with the number. If all data are in the paper and supplementary materials, include the sentence “All data are available in the main text or the supplementary materials.”

**Table 1. Properties of example compositions.** Impact of different building blocks on the properties of the resulting elastomer for 4 examples and 4 comparative examples.

| | **Composition** | | | | **Properties** | | |
|---|---|---|---|---|---|---|---|
| | Block C: Vinyl-PDMS/ PDMS-H | Block A: Vinyl-PDMS | Block B: PDMS-H | Block D: UV-PC | Optical Transmittance(%) | Tensile Strength (MPa) | Tensile Elongation (%) |
| Ex1 | 40 g | 0 g (0% w/w) | 0 g (0% w/w) | 40 µL (100 ppm) | 99 | 1.5 | 650 |
| Ex2 | 40 g | 0 g (0% w/w) | 0 g (0 % w/w) | 120 µL (300 ppm) | 99 | 3.3 | 500 |
| Ex3 | 40 g | 0 g (0 % w/w) | 0 g (0 % w/w) | 200 µL (500 ppm) | 96 | 3.0 | 620 |
| Ex4 | 40 g | 0 g (0 % w/w) | 0 g (0 % w/w) | 400 µL (1000 ppm) | 91 | 2.7 | 670 |
| CEx 1 | 40 g | 3 g (7.5 % w/w) | 0 g (0% w/w) | 129 µL (300 ppm) | 99 | 3.1 | 660 |
| CEx 2 | 40 g | 6 g (15 % w/w) | 0 g (0 % w/w) | 129 µL (300 ppm) | 99 | 2.3 | 900 |
| CEx 3 | 40 g | 0 g (0% w/w) | 3 g (7.5 % w/w) | 138 µL (300 ppm) | 99 | 4.1 | 290 |
| CEx 4 | 40 g | 0 g (15% w/w) | 6 g (15% w/w) | 138 µL (300 ppm) | 99 | 4.4 | 200 |

# Figures

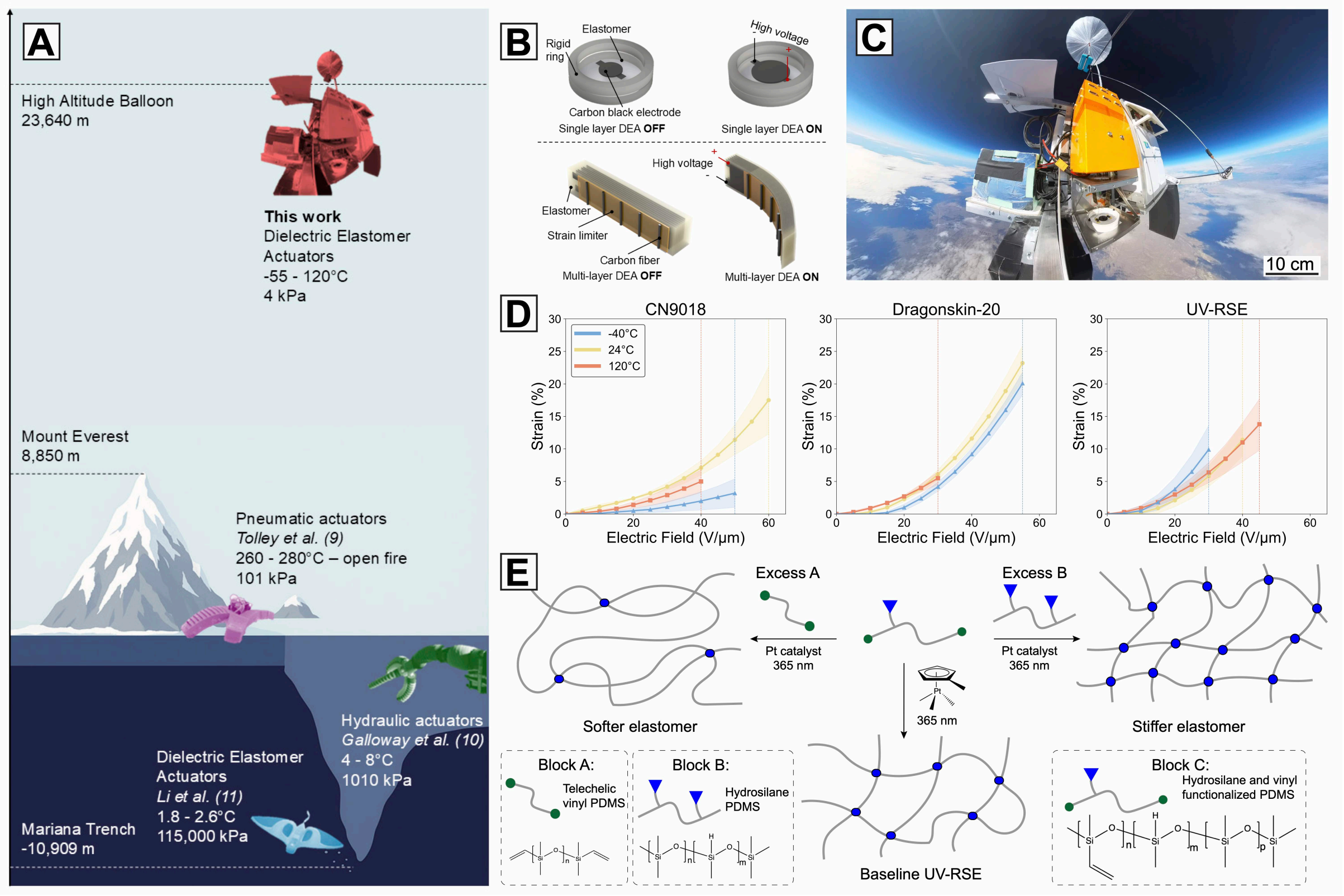

**Fig. 1. Demonstrations in extreme environments and key building blocks.** (A) Diagram of example soft actuators demonstrated at high temperatures and pressures *(9-11)*, compared to our stratospheric demonstration. (B) Schematic of single and multi layer dielectric elastomer actuators in their ON and OFF states. (C) Wide angle picture of the first stratospheric flight, showing a single layer actuator with two multilayer fingers as part of a gripper, recorded around 16.7 km elevation. (D) Electro-mechanical strain as a function of the applied electric field for an acrylic elastomer (CN9018 *(19)*), a commercial silicone elastomer (Dragonskin-20) and the novel formulation UV-curable resilient silicone elastomer (UV-RSE) at three temperatures (-40°, 24° and 120° C). (E) Chemical building blocks for the UV-RSE, and their impact on the overall elastomer network, all cured under UV light in the presence of a platinum catalyst.

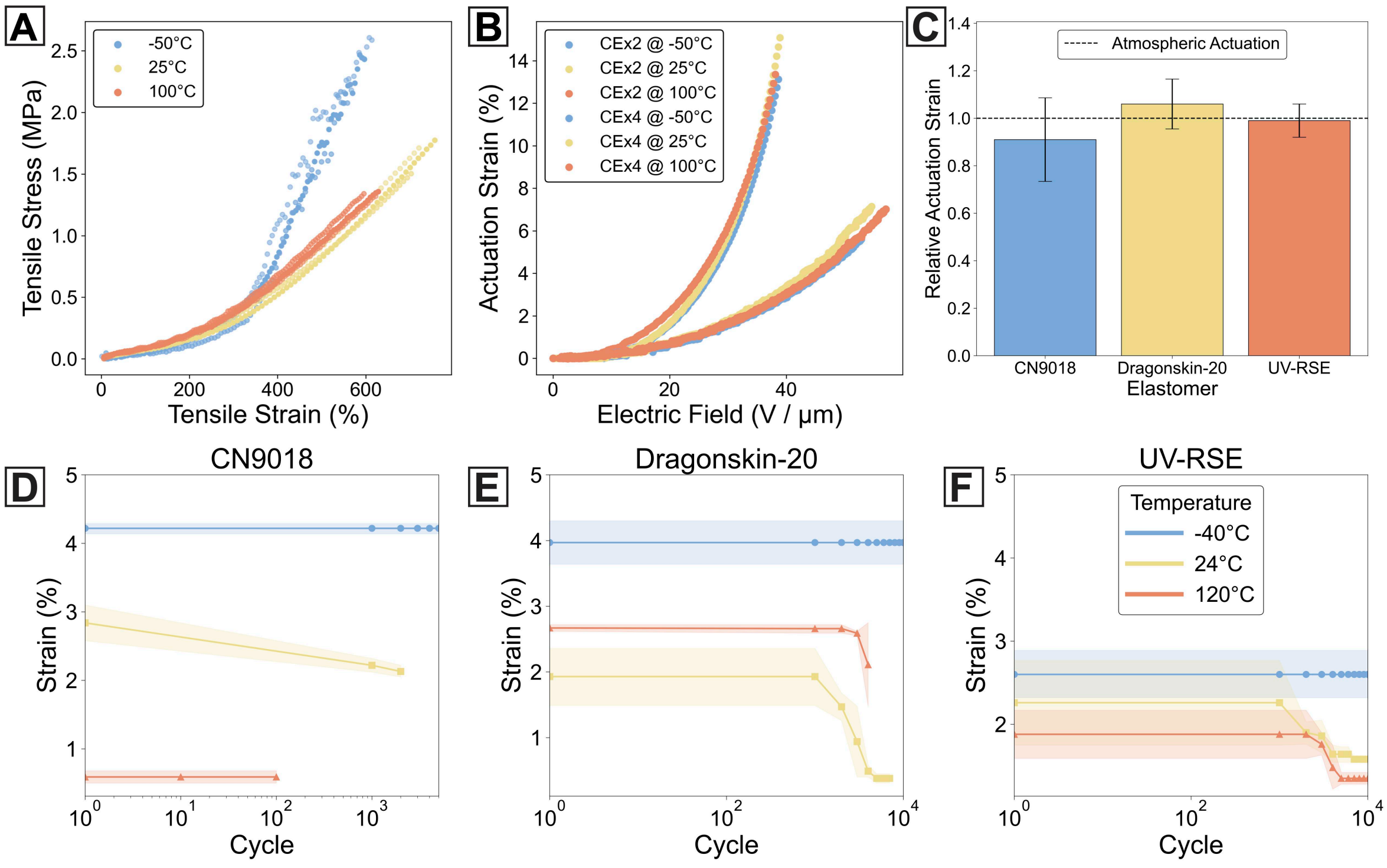

**Fig. 2. Thermal and pressure resilience.** (A) Stress vs. strain curve for a soft elastomer (CEx 2, with excess Block A) at -50°, 25° and 100 C°. (B) Electro-mechanical strain as a function of applied electric field for two elastomers (CEx2 - softer material made with excess Block A, and CEx4 - stiffer material made with excess Block B) at -50°, 25° and 100 C°. (C) Variation in the actuation strain for ten actuator samples each for the three elastomer types (acrylic CN9018, commercial silicone Dragonskin-20, and the UV-RSE made with the Ex2 composition), tested at 4 kPa relative to their performance at atmospheric pressure. (D) Cycle life evaluation for the same three elastomers under continuous 1 Hz actuation to 50% of the average breakdown field for each material up to 10,000 cycles, at -40°, 25° and 120 C°.

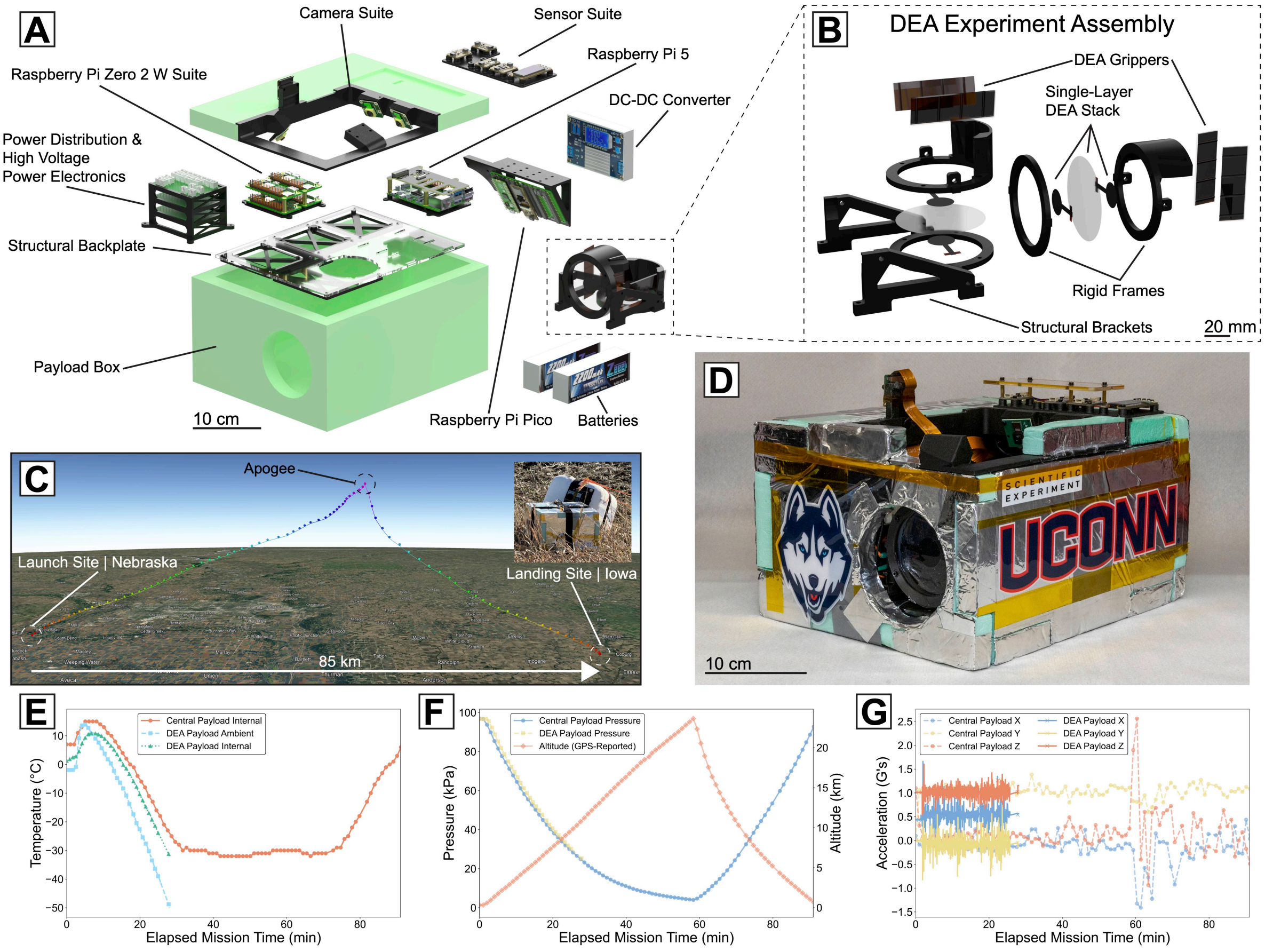

**Fig. 3. Payload and mission parameters.** (A) Computer aided design (CAD) of the payload from the second mission which showed actuator operation during flight. The payload composition is described including energy storage, power boosters, sensors and actuator payload.(B) Zoomed in view of the actuator payload, including the frames to support the two finger gripper components, and the ring supporting a single layer actuator. (C) Flight path of the mission, which traveled 85 km to an apogee of 23.6 km. (D) Photograph of the payload. (E) Recorded temperature for the flight showing the readings from enclosed sensors in the central payload, as well as the temperature recorded by sensors placed next to the actuators, showing a minimum temperature of -50° C before data loss. (F) Pressure and elevation during the mission, showing data loss occurred around 12 km of elevation. (G) Accelerometer data from both the central payload and our sensors showing the average G-forces experienced during the mission.

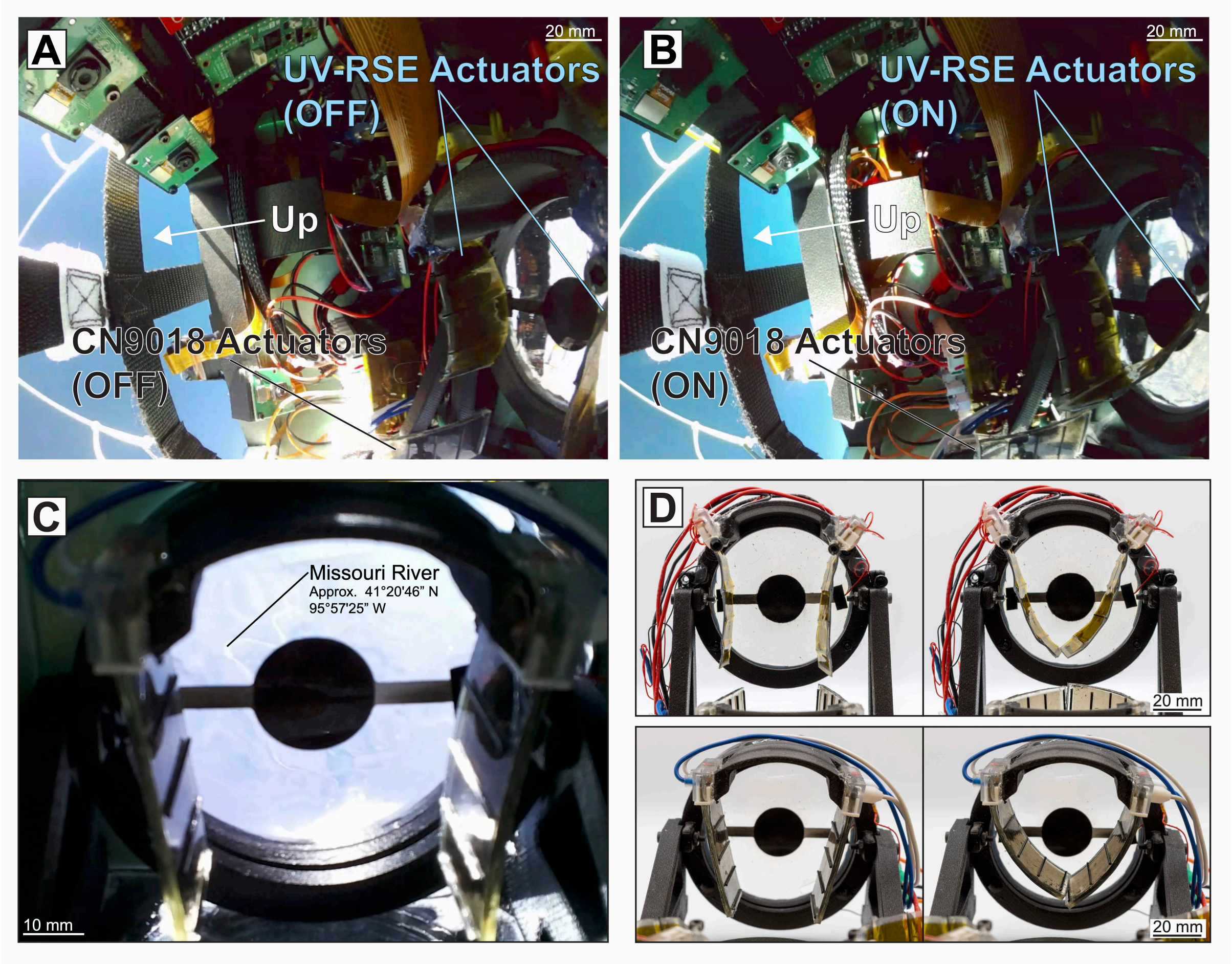

**Fig. 4. Operation at elevation and survivability.** (A) Photos of actuator in operation before data loss at 12 km, showing UV-RSE actuators operational while the CN9018 actuators are not responding to an electric field. (B) In flight photograph of a geographic landmark, the Missouri River, through the UV-RSE elastomer used for one of the single layer actuators. (C) Soft finger gripper actuators tested after the payload was retrieved, showing both sets of actuators (CN9018 and UV-RSE) retain full actuator function after the completion of the mission.